\documentclass[acmsmall]{acmart}
\usepackage{amsthm}
\usepackage{bm}
\usepackage{cleveref}
\usepackage{color}
\usepackage[inline]{enumitem}
\usepackage{float}
\usepackage{mathtools}
\usepackage{multirow}
\usepackage{siunitx}
\usepackage{subcaption}
\usepackage{graphbox}
\usepackage{tabularx} 

\ccsdesc[500]{Computing methodologies~Neural networks}
\ccsdesc[500]{Computing methodologies~Supervised learning by regression}
\ccsdesc[300]{Mathematics of computing~Multivariate statistics}
\keywords{Deep Learning, Multivariate Time Series Forecasting, Pruning, Transformer} 

\definecolor{haiblue}{HTML}{005AA0}
\definecolor{haigreen}{HTML}{8CB423}
\definecolor{haiyellow}{HTML}{FFD228}
\definecolor{haiorange}{HTML}{F0781E}
\definecolor{haired}{HTML}{D23264}
\definecolor{haidarkred}{HTML}{A0235A}
\definecolor{haicyan}{HTML}{50C8AA}

\newcommand{\Transformer}{\textcolor{haiblue}{Transformer}}
\newcommand{\Informer}{\textcolor{haigreen}{Informer}}
\newcommand{\Autoformer}{\textcolor{haiorange}{Autoformer}}
\newcommand{\FEDformer}{\textcolor{haicyan}{FEDformer}}
\newcommand{\Crossformer}{\textcolor{haidarkred}{Crossformer}}

\title[Pruning Time Series Transformers]{A Comparative Study of Pruning Methods in Transformer-based Time Series Forecasting}
\author{Nicholas Kiefer}
\email{nicholas.kiefer@kit.edu}
\orcid{0009-0008-1454-4144}
\affiliation{%
 \institution{Scientific Computing Center, Karlsruhe Institute of Technology}
 \city{Karlsruhe} 
 \country{Germany}
 \postcode{76137} 
}
\author{Arvid Weyrauch}
\email{arvid.weyrauch@kit.edu}
\orcid{0000-0002-2684-0927}
\affiliation{%
\institution{Scientific Computing Center, Karlsruhe Institute of Technology}
 \city{Karlsruhe} 
 \country{Germany}
 \postcode{76137} 
}
\author{Muhammed Öz}
\email{muhammed.oez@kit.edu}
\orcid{0009-0003-4337-407X}
\affiliation{%
\institution{Scientific Computing Center, Karlsruhe Institute of Technology}
 \city{Karlsruhe} 
 \country{Germany}
 \postcode{76137} 
}
\author{Achim Streit}
\email{achim.streit@kit.edu}
\orcid{0000-0002-5065-469X}
\affiliation{%
\institution{Scientific Computing Center, Karlsruhe Institute of Technology}
 \city{Karlsruhe} 
 \country{Germany}
 \postcode{76137} 
}
\author{Markus Götz}
\email{markus.goetz@kit.edu}
\orcid{0000-0002-2233-1041}
\affiliation{%
\institution{Scientific Computing Center, Karlsruhe Institute of Technology}
 \city{Karlsruhe} 
 \country{Germany}
 \postcode{76344} 
}
\affiliation{
 \institution{Helmholtz AI}
\city{Eggenstein-Leopoldshafen} 
 \country{Germany}
 }
\author{Charlotte Debus}
\email{charlotte.debus@kit.edu}
\orcid{0000-0002-7156-2022}
\affiliation{%
\institution{Scientific Computing Center, Karlsruhe Institute of Technology}
 \city{Karlsruhe} 
 \country{Germany}
 \postcode{76137} 
}

\begin{abstract}
The current landscape in time-series forecasting is dominated by Transformer-based models. Their high parameter count and corresponding demand in computational resources pose a challenge to real-world deployment, especially for commercial and scientific applications with low-power embedded devices. Pruning is an established approach to reduce neural network parameter count and save compute. However, the implications and benefits of pruning Transformer-based models for time series forecasting are largely unknown. To close this gap, we provide a comparative benchmark study by evaluating unstructured and structured pruning on various state-of-the-art multivariate time series models. We study the effects of these pruning strategies on model predictive performance and computational aspects like model size, operations, and inference time. Our results show that certain models can be pruned even up to high sparsity levels, outperforming their dense counterpart. However, fine-tuning pruned models is necessary. Furthermore, we demonstrate that even with corresponding hardware and software support, structured pruning is unable to provide significant time savings.

\end{abstract}

\begin{document}

\maketitle

\section{Introduction}

Time series analysis and forecasting is a central task in many commercial and scientific applications, such as meteorology, particle physics, life sciences, and energy research.
Examples include the forecasting of electricity generation and consumption~\cite{electricity_Vicente, recycle}, weather phenomena~\cite{nguyen2023scalingtransformerneuralnetworks} or technical components in particle detectors~\cite{Petnet}.
While deep learning techniques have an established position next to classical statistical methods for quite some time, the introduction of the Transformer architecture~\cite{vaswani_attention_2023} for natural language processing (NLP) and its adaptation to time series data have truly revolutionized the field~\cite{zhou_informer_2021, zhou_fedformer_2022, wu_autoformer_2022, liu2024itransformer}.
Transformers are, however, known to be computationally demanding, both in memory and training time, due to their large parameter count, the quadratic complexity with respect to the sequence length, and their heavy use of dense layers with large hidden dimensions~\cite{fournier_practical_2023}. Indeed, experiments on scaling laws have demonstrated that linear increases in predictive capabilities of Transformer models require exponential increases in trainable weights~\cite{kaplan2020scalinglawsneurallanguage}, which results in a growing need for computational resources.

Similarly, Transformer-based models for time series forecasting problems are growing with respect to their parameter counts, as illustrated in \Cref{tab:model_size}. Additionally, these models quickly suffer from an overfitting problem, due to an imbalance between parameter count and training dataset size~\cite{Zeng_Chen_Zhang_Xu_2023, jin2024llmbasedknowledgepruningtime}.

\begin{table}
    \centering
    \caption{Recent proposed Transformer-based models for time series forecasting task, their parameter count and publication year.}
    \label{tab:model_size}
    
    \begin{tabularx}{0.8\linewidth}{X l c c} 
    \toprule
    \textbf{Model} & \multicolumn{1}{c}{\textbf{Characteristic}} &\textbf{Year} & \textbf{Params (M)} \\
    \midrule
    Transformer & Language translation & 2017 & 10 \\
    Informer & Probabilistic attention & 2020 & 11 \\
    Autoformer & Sequence decomposition & 2020 & 10 \\
    FEDformer & Fourier transform & 2022 & 16 \\
    Time-LLM & LLM capabilities & 2024 & 3405 \\
    \bottomrule
    \end{tabularx}
\end{table}

This high demand in memory and computation time of Transformers stands in sharp contrast to their prospective deployment on low-resource devices, as required in real-world applications~\cite{chen2024personalizedadapterlargemeteorology}.

Meanwhile pruning, i.e., removal of neural network weights estimated to be unnecessary for the predictive performance, has become an established method to reduce computational demand, in particular for model deployment~\cite{tprune}. 
Pruning is based on the observation that the majority of weights in a neural network are close to zero after training, and thus, do not contribute to the output~\cite{optimal_brain_damage_lecun}. 
Consequently, these weights can be removed from the neural network weight matrices, thereby reducing the number of numerical operations performed in every pass of the model. The corresponding networks are termed \emph{sparse}.

Since the inception of sparse neural networks, research interest in network pruning has spiked, and numerous methods and algorithms have been proposed in the last five years~\cite{hoefler_sparsity_2021}. While the natural occurrence of sparsity and the phenomenon of an a-priori better sparse networks, so-called lottery tickets~\cite{frankle_lottery_2019}, has only been observed in selected models, it is generally assumed that these findings translate across network architectures and learning tasks~\cite{liu2024surveylotterytickethypothesis}. 

Pruning of Transformer-based models for time series forecasting is a promising approach to reduce computational resource demand in real-world application, but has not been studied to date to the best of the authors' knowledge. 
In this work, we aim to close this gap by providing a corresponding benchmark study. We empirically study the behavior of previously proposed time series models under the effect of pruning with respect to dataset and model size.
In summary, our contributions are:
\begin{enumerate}
    \item We train and prune five state-of-the art Transformer-based models for time series forecasting on a number of datasets, using both unstructured and structured pruning. We investigate different data set sizes, and apply models both in their originally published version and with adapted smaller parameter count, to rule out the effect of overfitting.
    \item We then evaluate model performance with respect to the loss on the test dataset, the number of non-zero parameters, the parameter density, and the number of floating point operations (FLOPs). We further investigate, how fine-tuning after pruning can aid in recovering predictive performance lost during the pruning phase.
\end{enumerate}

Our findings provide insight into the benefits and drawbacks of pruning Transformer-based models for time series forecasting in terms of predictive performance and resource requirements.


\section{Related Work}
\label{sec:related_work}

\paragraph{Transformer-based Time Series Forecasting}
Multivariate time series (MTS) forecasting is the task of forecasting the next $T_{pred}$ time steps of a given times series $x$ with length $T_{inp}$, where each time step has a real-valued feature vectors with dimension $D$. These $D$ univariate series are assumed to be dependent and relevant to the task. Models for MTS forecasting have to tackle multiple problems: dependencies and patterns can emerge inside a single dimensional long time series, and so the models have to be adequate to handle long sequences without forgetting; events may be correlated across time and across features.

Deep learning-based MTS forecasting has received increasing attention since the introduction of the Transformer~\cite{vaswani_attention_2023}. Originally introduced for NLP, the ability to process and forecast sequential data makes it a natural fit for temporally resolved data. 
However, while the capability of the self-attention-mechanism to relate individual sequence elements to one another provides impressive predictive skills, the $\mathcal{O}(L_{inp}L_{out})$ complexity in sequence length $L$ poses computational challenges, especially for longer time series.
Hence, researchers have dedicated special attention to reducing this computational complexity with respect to memory and computational load~\cite{fournier_practical_2023}.

One of the first Transformer-based models specifically designed for time series forecasting was the \textbf{Informer} model~\cite{zhou_informer_2021}. It introduced the \emph{ProbSparse} attention mechanism, which attends to the most important of a random subset of past time series tokens, thereby achieving a theoretical complexity of $\mathcal{O}(n\log{}n)$.
The \textbf{Autoformer}~\cite{wu_autoformer_2022} achieves the same complexity as the Informer by introducing a decomposition architecture with an auto-correlation mechanism. This feature specifically addresses the periodicity of time series data. The \textbf{FEDformer} model~\cite{zhou_fedformer_2022} is a frequency-enhanced Transformer with a complexity of $\mathcal{O}(n)$. It expands upon the frequency spectrum idea, performing attention in the Fourier domain. While the above mentioned models are designed for multivariate time series, the \textbf{Crossformer}~\cite{zhang_crossformer_2022} was proposed specifically to capture cross-dimension dependencies with an $\mathcal{O}(DS^2)$ memory complexity. The complexity is a function of data dimensions $D$ and the segment size $S$, instead of the whole sequence length $L$. The sequence is split up in a pre-defined number of segments.
Besides architectural advances, other approaches have focused on improving tasks like classification or anomaly detection in time series~\cite{wen_transformers_2023}.

\paragraph{Pruning}
Neural network pruning has a long-standing history ever since its inception. For a general survey with a broad overview about pruning techniques and taxonomy, see~\cite{cheng_survey_2023}. 

Pruning has various purposes, such as reducing the model size and thereby computational demands, increasing inference speed or as a form of regularization. 
Existing techniques can be largely categorized into \textbf{unstructured} and \textbf{structured} pruning. 
Unstructured pruning of parameters is the method of zeroing out entries that are estimated to be unimportant to the networks performance. A straight-forward approach is to remove the parameters with the lowest Minkowski-norm, given a certain threshold~\cite{hagiwaraRemoval}.
Without the usage of specialized hardware or software, unstructured pruning amounts to masking out the pruned model weights, and thus, does not achieve actual speedup or computational savings~\cite{cheng2023surveydeepneuralnetwork, 3DSemanticSegmentationWithSubmanifoldSparseConvNet}. It is nevertheless easy to implement and viable on a wide range of networks, delivering theoretical insights for more compact and performant networks.

Structured pruning refers to pruning groups of parameters. For example, in linear layers such a group could be all weights connecting to a single neuron, i.e., a node. In that case, pruning amounts to removing whole rows or columns in the weight matrix. Other structured pruning mechanisms include for example filter pruning in CNNs. Indeed, it has been demonstrated that the structural changes lead to an actual speed-up, providing a valuable benefit for applications~\cite{liang_pruning_2021}.
For a detailed survey of structured pruning techniques, see \cite{he_structured_2024}. 
While pruning a simple multi-layer perceptron (MLP) is easy, the introduction of skip-connections in residual networks~\cite{he2015deepresiduallearningimage} and multiple attention-heads in Transformers~\cite{vaswani_attention_2023} requires additional record keeping and adjustments. \emph{DepGraph}~\cite{fangDepGraphAnyStructural2023} provides a technique to prune a model based on a constructed dependency graph that groups parameters based on their connection in the model's forward pass. 

The most common method is to prune a fully trained network, possibly with fine-tuning after the pruning step. Pruning after training has been shown to not affect final performance significantly, in some cases even improving it~\cite{frankle_lottery_2019, oz2024model}. Furthermore, pruning at different stages of the training has been investigated~\cite{nowak_fantastic_2023}. Pruning-at-initialization and pruning-and-regrowing, for example, set the pruning phase before or during training~\cite{wang_recent_2022}. 

Because of their high computational cost and parameter count, Transformers are the main study subject of current pruning research. Recent state-of-the-art models will not fit on a single GPU during training, with forecasts for Transformers showing that increasing model size is still desirable~\cite{kaplan_scaling_2020}. Although parameters of Transformers are mostly found in linear layers, which can be targeted by weight, node and layer pruning, the multi-head architecture also allows for structured head pruning~\cite{behnke_losing_2020}. A pruning approach for Transformers on a module level has been investigated by~\cite{wang_enhancing_2022}, where modules are building blocks (input embedding, multi-head attention, batch normalization, etc.) of the Transformer. An orthogonal method is token pruning, as proposed in~\cite{bolya_token_2023}. The pruning of Transformers for NLP has been covered extensively~\cite{chitty-venkata_survey_2023, fournier_practical_2023, tang_survey_2024}; similar research exists for the Vision Transformer~\cite{papa_survey_2024, chen_comprehensive_2024}. This leaves the field of time series forecasting open to pruning experiments on proposed Transformer-based models.

\section{Experimental Setup}
\label{sec:experiments}

In the following, we provide an in-depth experimental benchmark on pruning Transformer-based models for multivariate time series forecasting. We train several state-of-the-art models on a number of commonly used datasets. These trained models are then pruned to increasing levels of sparsity, using unstructured magnitude pruning as well as structured pruning with \emph{DepGraph}~\cite{fangDepGraphAnyStructural2023}. The pruned models are evaluated with respect to their density, predictive performance and, in case of structured pruning, computational speedup and FLOPs. We further fine-tune the best pruned model in each category, to make-up for any loss in predictive performance originating from pruning, and evaluate the corresponding models as well.

\subsection{Models and Training}
\label{subsec:exp-models}

For our benchmark study, we use five state-of-the art Transformer-based models for MTS forecasting: the vanilla Transformer~\cite{vaswani_attention_2023}, Informer~\cite{zhou_informer_2021}, Autoformer~\cite{wu_autoformer_2022}, FEDformer~\cite{zhou_fedformer_2022} and Crossformer~\cite{zhang_crossformer_2022}.
We train these models on different datasets, described in~\cref{subsec:exp-data}, with forecasting horizons of $T_{pred} = 96, 192, 336, 720$. 
Hyperparameters for all model trainings are set following the publications of FEDformer and Crossformer: all models are trained on all datasets and forecast lengths using the Adam~\cite{kingma2017adammethodstochasticoptimization} optimizer, a batch size $B=32$, learning rate $\eta=5\mathrm{e}{-4}$.
For naive pruning experiments (c.f. ~\cref{subsec:exp-pruning}), model architectures are utilized in their originally published version, with two encoder and one decoder layers, eight heads, and a dropout rate of $0.2$. To rule-out overfitting effect of these large models on the comparably small datasets, we further conduct experiments with reduced model sizes (\cref{subsec:exp-smallmodels}).
The training loss function is the mean-square error (MSE). As done by the authors of FEDformer, we train all models for ten epochs with an early stop with patience three.
For reproducibility and statistics, all models are trained three times for every experiment, i.e., dataset and forecasting horizon combination. 

All experiments were run on a compute node equipped with 12 Intel Xeon Platinum 8368 CPUs and four NVIDIA A100 or H100 GPUs as accelerators with 40GB and 96GB VRAM, respectively. Code was implemented using Python 3.9, PyTorch 2.2, CUDA 12.1 and torch-pruning 1.3. All code is publicly available under \footnote{\textit{Will be made publicly available upon publication}}.

\subsection{Datasets}
\label{subsec:exp-data}

We perform experiments on well-known time series datasets~\cite{liu2022pyraformer, du2022preformerpredictivetransformermultiscale, wen2023transformerstimeseriessurvey} including the following: ETT, ECL, Exchange, Traffic, and Weather, made available by Autoformer~\cite{wu_autoformer_2022} and Informer~\cite{zhou_informer_2021}. 
ETT includes four temperature datasets collected from two electricity transformers, with two different time resolutions, consequently named ETTm1, ETTm2, ETTh1, ETTh2. Due to brevity we will show results for ETTm1 only.
ECL contains the electricity consumption of 321 different costumers from 2012 to 2014.
The Exchange dataset is a record of the daily exchange rates of eight countries from 1990 to 2016.
Traffic is a dataset comprising the occupation rate of freeways in the San Francisco Bay area.
The Weather dataset is the set of 21 meteorological indicators recorded by stations in Germany, covering one year.
For experiments on training and pruning models on large dataset sizes (c.f. \cref{subsec:exp-largedata}), we compile a dataset from the Transparency Platform~\cite{entsoe}, operated by the European Network of Transmission System Operators for Electricity (ENTSO-E). We will name this the ENTSO-E dataset. It contains hourly electricity grid data of 56 zones in Europe, like load, generation power flow, etc. starting from 2015. 

Details on the number of features and timesteps (samples) are listed in Table~\ref{tab:datasets}.
Datasets are each split into training, validation and testing set by a ratio of 7:1:2.

\begin{table}
    \centering
    \caption{Dataset statistics, summarizing the feature dimensions, the number of data points available, and sampling frequency with which the data was recorded.}
    \label{tab:datasets}
    
    \begin{tabularx}{0.8\linewidth}{X r r r} 
    \toprule
    \textbf{Dataset} & \textbf{Features} & \textbf{Samples} & \textbf{Granularity} \\
    \midrule
    ETT & 7 & 17,421 & 1 \unit{\hour} / 5 \unit{\minute}\\
    ECL & 321 & 26,305 & 1 \unit{\hour}\\
    Exchange &  8 & 7,598 & 1 \unit{\day}\\
    Traffic & 862 & 17,545 & 1 \unit{\hour}\\
    Weather & 21 & 52,697 & 10 \unit{\minute}\\
    ENTSO-E & 1 & 800,412 & 1 \unit{\hour}\\
    \bottomrule
    \end{tabularx}
\end{table}

\subsection{Pruning}
\label{subsec:exp-pruning}

The \emph{unstructured} weight pruning sets the smallest parameters to zero, resulting in a sparsity $s$, a fractional number denoting the number of zeros in relation to full parameter count, defined as 
\begin{equation}
    s = \frac{\text{number of zeros}}{\text{number of trainable parameters}}
\end{equation}.
The parameter density $d$ is then defined as the complement of the sparsity, $d = 1 - s$.
We define ten target sparsity levels, spread uniformly in log-space
\begin{equation}
    s = 1 - 0.8^i, i=0,...,10
\end{equation}.
A fixed pruning ratio allows us to compare how models fare with an equal ratio of their original parameters. However, this does not imply that all models have an equal parameter budget.

Current frameworks will only mask pruned elements and do not support unstructured sparsity any further, for example by employing specialized kernels for sparse matrix-vector-multiplication (SpMV)~\cite{gao2024systematic, paszke2019pytorchimperativestylehighperformance}. This mask is calculated with the PyTorch library~\cite{pytorch}. It has the same shape as all weight matrices, and its binary elements indicate pruned weights. For inference, this mask can be applied to the original matrix, setting corresponding elements to zero.

On the other hand, \emph{structured} weight pruning is able to yield performance gains by removing rows of the weight matrix, which reduces the computational load in the matrix-vector-multiplication. 
For structured weight pruning, we utilize the pruner of the \textit{torch-pruning} framework~\cite{fangDepGraphAnyStructural2023}. A group of parameters is defined as in \emph{DepGraph}~\cite{fangDepGraphAnyStructural2023}. We set the pruner to the same pruning ratios as for unstructured pruning. Note that in this case, a set pruning ratio is not necessarily achieved due to dependencies in the graph.

\subsection{Fine-tuning}
\label{subsec:exp-finetune}

To make up for any losses in predictive accuracy incurred during the pruning phase, it is common to add a second training phase after pruning, following the scheme train-prune-fine-tune. For the cost of additional compute resources, one can attain a higher sparsity rate with equal predictive performance. We performed fine-tuning on each of the pruned models trained on the ETTm1 dataset with a forecast length of $192$, as this is representative of a medium-length forecast. We load the pruned model and continue training with the same setup as in the training phase, either for ten epochs or early stopping with a patience of three. In the unstructured pruning setup, we retain the sparsity level by applying the corresponding mask in every update step.

\subsection{Evaluation Metrics}
\label{subsec:exp-metrics}

After pruning, the models are evaluated on the hold-out test data and compared to their unpruned version with respect to the following metrics.

For quantifying predictive performance of the models, the MSE training loss function is equally used for evaluation on the validation and testing dataset. It is defined as
\begin{equation}
    \text{MSE} = \frac{1}{n} \sum_{i=1}^{n} (y_i - \hat{y_i}) ^ 2
\end{equation}
with $y_i$ the ground truth and $\hat{y_i}$ the output of the model. 

Further, the count of a models parameters is the sum of all trainable weight matrices. Runtime measurements are done by measuring the time between CUDA events before and after an example batch of data is put through the model 500 times, with 50 warm-up steps. 

In case of structured pruning, inference time speedup is compared to the unpruned model. We compute speedup by dividing two measurements of a model, unpruned and pruned. The resulting factor is an indicator how much faster a pruned model is and quantifies the computational performance.

Similarly, the Floating Point Operation (FLOP) reduction is calculated by dividing the FLOP count of the unpruned model by the FLOP count of the pruned one.

\subsection{Reducing model size to rule-out overfitting}
\label{subsec:exp-smallmodels}

When training models in the previous experiments, we observe very fast model convergence, hinting that the models in their originally published version are too large, and thus run into overfitting. Overfitting models on small datasets with little to no regularization is a valid concern, especially when smaller models could yield the same performance, at a lower computational cost. To investigate the effects of pruning in smaller model sizes and rule-out the effect of overfitting and loss of predictive performance, we additionally train model instances with linear embedding layer sizes scaled down by a factor of 10 on the ETTm2 dataset (corresponding to our highest pruning rate), thereby significantly reducing the number of trainable parameters (see \cref{tab:small_vs_big}). Apart from the model size, we keep an equal training setting compared to the big models. 

\subsection{Increasing dataset size to rule-out overfitting}
\label{subsec:exp-largedata}

Overfitting can be induced either if the model is too large or if the dataset is too small. Hence, orthogonal to the previous experiment, we examine the performance of all models on a significantly larger dataset in terms of number of samples, features, diversity and general periodicity. For this, we utilize the ENTSO-E dataset. We train and evaluate the models on a representative slice of the load subset, zeroing out missing values. We perform 1.6 million gradient update steps, equivalent to two epochs. The models are then pruned through unstructured pruning as described above and evaluated on the test dataset. For comparison, appropriate smaller model sizes are chosen, trained and tested. Their size is determined by the pruning results, choosing the minimal size where performance is not impacted by parameter loss.

\section{Results}
\label{sec:results}
In this section we present the results of our experiments, starting with the unstructured pruning experiments. This is followed by the structured pruning results, the measured inference speedup from it, and fine-tuning results for unstructured pruning. We then present results on reducing model size and increasing dataset size.

\subsection{Unstructured weight magnitude pruning}

\begin{figure}
    \centering
    \includegraphics[width=.95\linewidth]{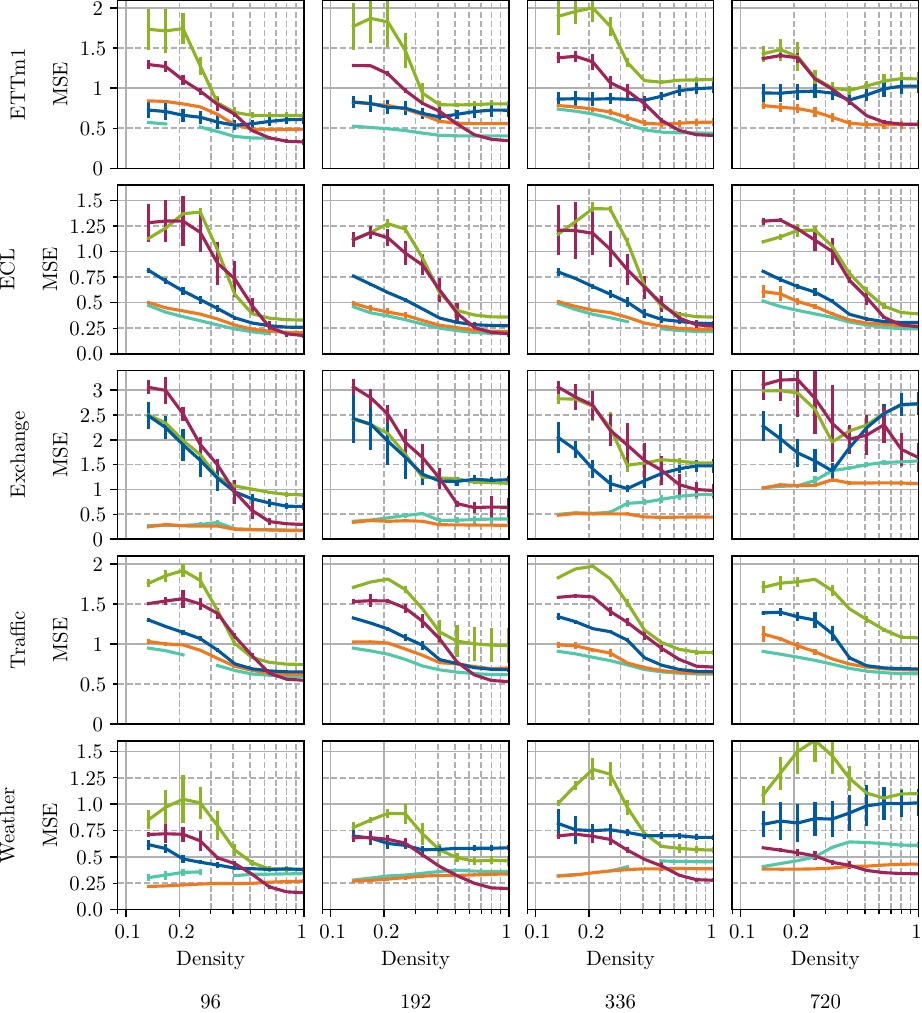}
    \caption{Pruning results for weight magnitude pruning. Plotted is the MSE on the test dataset over the parameter density of the models for all datasets and forecast lengths, with logarithmic scaling on the x-axis. Models are \Transformer, \Informer, \Autoformer, \FEDformer, and \Crossformer. Best viewed zoomed in.}
    \label{fig:nn_prune}
    \Description{An image of a grid of plots showing the impact of unstructured weight magnitude pruning on the performance of Transformer models across the datasets. The x-axis represents the density, from 0.1 to 1., while the y-axis shows the MSE as forecasting performance (both log-spaced). Each row corresponds to a different dataset (ETTm1, ECL, Exchange, Traffic, and Weather), and each column represents different forecast horizons (96, 192, 336, and 720 time steps).}
\end{figure}

Figure \ref{fig:nn_prune} shows the results of unstructured weight magnitude pruning. We report the average loss on the test dataset given the models pruned to predetermined sparsities. Missing datapoints in the plot are induced by models either diverging in training and producing exploding gradients, or out-of-memory errors during training, especially on the longest forecasting horizon. We observe the FEDformer to be the most unstable.
As assumed, all models perform worse with a vanishing number of parameters, as they lose predictive performance, indicated by the higher loss on the test dataset. The strength of this effect, however, varies between models, datasets and, on a lesser note, forecast length.

We observe that nearly all models can be pruned to around 50\% sparsity without any significant performance loss.
The pruned versions of the Informer retain the least performance through pruning; in many cases they regain a lower loss through further pruning, on ETTm1, ECL, Traffic and Weather. This is likely due to regularization effects, which are lost again at high sparsities.
The Autoformer achieves the lowest loss throughout, i.e., the best predictive skill, together with the FEDformer. Even pruning the Auto- and FEDformer models to 1\% of their original parameter count does not yield a higher loss than other models with a lower sparsity. The instability of the FEDformer during training makes the Autoformer the best choice for consistent training and pruning behaviour.

\subsection{Structured node pruning}
\begin{figure}
    \centering
    \includegraphics[width=.95\linewidth]{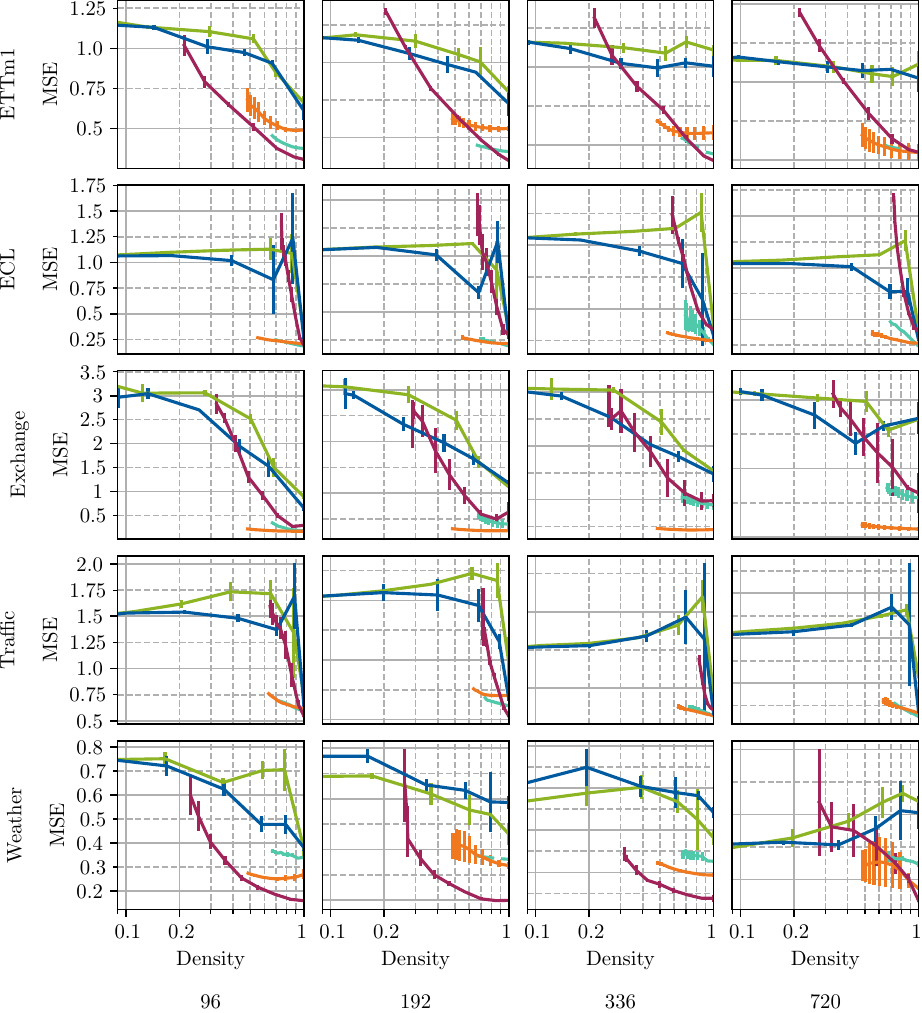}
    \caption{Pruning results for structured node pruning using \textit{torch-pruning}. Plotted is the MSE on the test dataset over the measured parameter density for all forecast lengths and datasets. Models are \Transformer, \Informer, \Autoformer, \FEDformer, and \Crossformer. Best viewed zoomed in.}
    \label{fig:tp_prune}
    \Description{An image of a grid of plots showing the impact of structured node pruning on the performance of Transformer models across all datasets. The x-axis represents the density from 0.01 to 1., while the y-axis represents the MSE as forecasting performance (both log-spaced). Each row corresponds to a different dataset (ETTm1, ECL, Exchange, Traffic, and Weather), and each column represents different forecast horizons (96, 192, 336, and 720 time steps).}
\end{figure}

We present the results of pruning the models with the \emph{DepGraph} pruner from \textit{torch-pruning} in Figure \ref{fig:tp_prune}. We observe, that
while the vanilla Transformer and the Informer can be pruned to 1\% of their original parameter count, the pruning algorithm fails to prune Autoformer and FEDformer to the desired level of sparsity. Furthermore, the pruner fails to reach very high sparsities, instead, a lower sparsity is reproduced, e.g., the plotted line for these models has a "turning point".
The Transformer, Informer and Crossformer lose predictive performance immediately on the ECL and Traffic dataset.
Overall, the Autoformers performance does not change much, except for high sparsities on the ETTm1 dataset for the short $96$ step forecast length. Minor constant losses can be found for the FEDformer on all datasets.
For the longest forecast length some models have a lower loss when pruned, e.g., the Transformer and Informer on the Weather dataset.
On the Exchange dataset, all models lose predictive performance uniformly. It is clear that the \emph{DepGraph} pruner fails to achieve a consistent pruning behavior across models and datasets, making its node-pruning algorithm only viable in certain model-dataset combinations.

\subsection{Inference time speedup}

Models trained on the ETTm1 dataset, for a forecast length of 192, and structurally pruned to a density of $d=0.8^5 \approx 0.33$ are evaluated for their actual density, speedup compared to the dense counterpart and MSE in Table \ref{tab:speed}. We do not find a significant speedup except for the Informer model, which is also the most pruned model with a remaining density of 11\%. Contrarily, The FEDformer is slower than unpruned, even though it is pruned to a density of 74\%, indicating that the CUDA kernels can not take advantage of the smaller weight matrices.

Compared to the unpruned version, every model has a reduced FLOP count, up to seven times (for the Informer) smaller. Both the FEDformer and Autoformer have the least reduction with a factor of $<2$.

\begin{table}
    \centering
    \caption{Inference time speedup (unpruned batch inference time over pruned time), MSE and measured density, when setting target pruning ratio to 33\% density in \textit{torch-pruning} and evaluating on the testing dataset.}
    \label{tab:speed}
    
    \begin{tabularx}{\linewidth}{X r r r r}
    \toprule
    \textbf{Model} & \textbf{Density} & \textbf{FLOP reduction}$\uparrow$ & \textbf{Speedup$\uparrow$} & \textbf{MSE $\downarrow$} \\
    \midrule
    Transformer & 0.2182 & 4.43 & 1.21 & 1.1496 \\
    Informer & 0.1121 & 7.63 & 1.51 & 1.1249 \\
    Autoformer & 0.5123 & 1.97 & 1.31 & 0.5489 \\
    FEDformer & 0.7436 & 1.64 & 0.98 & 0.4346 \\
    Crossformer & 0.3396 & 3.44 & 1.22 & 1.0886 \\
    \bottomrule
    \end{tabularx}%
\end{table}

\subsection{Fine-tune after pruning}
The models trained on the ETTm1 dataset, for a forecast length of $192$, and weight magnitude pruned to a density of 33\% are trained again. The resulting predictive performance on the testing dataset is presented in table \Cref{tab:retraining}. The vanilla Transformer has better performance on the test set after pruning, which it loses again with fine-tuning. This discrepancy can likely be explained through overfitting on the training dataset and a regularizing effect of pruning. The Informer and Autoformer models have an approximately 20\% higher loss, which is not restored through fine-tuning. The FEDformer model has a lower loss after pruning and after fine-tuning. The Crossformer loses much of its predictive power with pruning, but is able to recover most of it through fine-tuning, to within 1\% of its original loss, making it the only model where fine-tuning seems necessary. Except for the Transformer, all models that were specifically proposed for time series forecasting have a greater or equal performance after fine-tuning. 

\begin{table}
    \centering
    \caption{MSE loss on the ETTm1 testing dataset, forecast length 192, after training, after pruning to 33\% and fine-tuning the pruned models. The difference between pruning and fine-tuning is also shown.}
    \label{tab:retraining}
    
    \begin{tabularx}{1.0\linewidth}{X r r r r}
    \toprule
    \textbf{Model}\textbackslash\textbf{MSE}$\downarrow$ & \textbf{Trained} & \textbf{Pruned} & \textbf{Fine-tuned} & \textbf{Difference} \\
    \midrule
    Transformer & 0.7256 & 0.6869 & 0.7749 & -0.0880 \\
    Informer & 0.8037 & 0.9393 & 0.8910 & 0.0483 \\
    Autoformer & 0.5591 & 0.6696 & 0.6031 & 0.0664 \\
    FEDformer & 0.5280 & 0.4403 & 0.4088 & 0.0314 \\
    Crossformer & 0.3456 & 0.8124 & 0.3502 & 0.4621 \\
    \bottomrule
    \end{tabularx}%
\end{table}

\subsection{Reducing model size to rule-out overfitting}
Cutting down on initial parameter size, models with reduced linear embedding layer size are trained on ETTm2, prediction length 192, to compare with their large counterparts. With a ten-fold reduction in total parameter count, all smaller models are able to deliver a lower loss score on the test dataset than their large counterparts, with precise results shown in table \Cref{tab:small_vs_big}. A notable exception is the Crossformer, which additionally experiences no change in time per epoch. We find in our experiments that the larger models yield lower training losses than the smaller ones, indicating overfitting on the training dataset, directly impacting performance on the test dataset not seen during training. The Transformer is improved the most by training a smaller model, with an MSE reduced by 38\%. Evidently, the ETTm2 dataset is already satisfactorily learned by models with small parameter count. Ideally, the model size has to be adjusted to the dataset.

\begin{table}
    \centering
    \caption{Total model parameter count (noted $\theta$), MSE test error and time in seconds/minutes per epoch for the ETTm2 and ENTSO-E dataset. Listed are both a large and small variant for each model. A dash indicates a failed experiment due to exploding gradients.}
    \begin{tabularx}{\linewidth}{l X r c r| r c r}
        \toprule
          & & \multicolumn{3}{c|}{\textit{ETTm2 (small)}} & \multicolumn{3}{c}{\textit{ENTSO-E (large)}}\\
         \textbf{Size} & \textbf{Model} & \textbf{Params (M)} & \textbf{MSE}$\downarrow$ & \textbf{t/ep (s)} & \textbf{Params (M)} & \textbf{MSE}$\downarrow$ & \textbf{t/ep (min)} \\
        \midrule
         \parbox[t]{2mm}{\multirow{5}{=}{\rotatebox[origin=c]{90}{Small}}} & Transformer & 0.37 & 0.3099 & 10.2 & 4.61 & 0.1810 & 3.1 \\
         & Informer    & 0.42 & 0.3364 & 7.6 & 5.40 &  0.2124 & 3.4\\
         & Autoformer  & 0.37 & 0.1914 & 10.0 & 4.61 &  0.4435 & 6.4 \\
         & FEDformer    & 1.81 & 0.2606 & 47.0 & 7.02 &  - & 22.1 \\
         & Crossformer & 1.18 & 0.4542 & 25.8 & 0.37 & 0.2111 & 16.7 \\
         \midrule
         \parbox[t]{2mm}{\multirow{5}{=}{\rotatebox[origin=c]{90}{Large}}} & Transformer  & 10.52 & 0.5020 & 29.2 & 10.52 &  0.1787 & 8.6 \\
         & Informer     & 11.31 & 0.4045 & 32.8 & 11.31 & 0.1814 & 8.6 \\
         & Autoformer   & 1.81 & 0.2049 &56.3 & 10.51 &  0.3885 & 18.0 \\
         & FEDformer    & 16.27 & 0.2806 & 134.1 & 16.27 &  - & 38.7 \\
         & Crossformer  & 11.63 & 0.4098 & 25.8 & 0.78 & 0.1978 & 15.7 \\
          \bottomrule
    \end{tabularx}
    \label{tab:small_vs_big}
\end{table}

\subsection{Increasing dataset size to rule-out overfitting}
\begin{figure}
    \centering
    \includegraphics[align=c, width=0.5\linewidth]{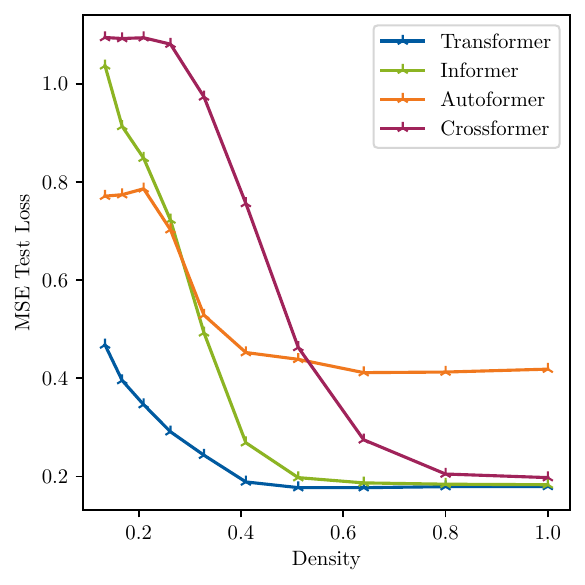}
    \caption{Pruning results for weight magnitude pruning on the ENTSO-E test dataset with prediction length 192. Plotted is the MSE of all large models over their parameter density. Models are \textcolor{haired}{Transformer}, \textcolor{haiorange}{Informer}, \textcolor{haigreen}{Autoformer}, \textcolor{haiblue}{FEDformer}, and \textcolor{haidarkred}{Crossformer}.}
    \label{fig:entsoe}
    \Description{An image of the weight magnitude pruning results after training on the ENTSO-E dataset. The x-axis is the density of the models going from 0.01 to 1.0, and the y-axis the MSE test error spanning from 0.2 to 1.0, with the Transformer, Informer, Autoformer and FEDformer plotted. }
\end{figure}

\begin{figure}
    \includegraphics[width=.5\linewidth]{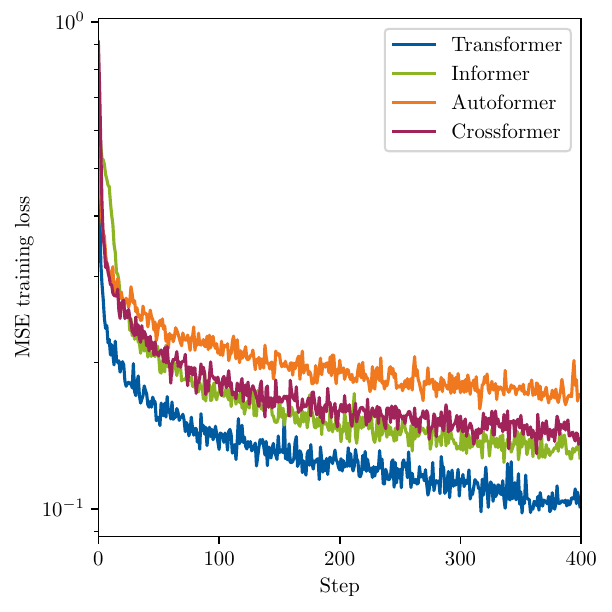}%
    \includegraphics[width=.5\linewidth]{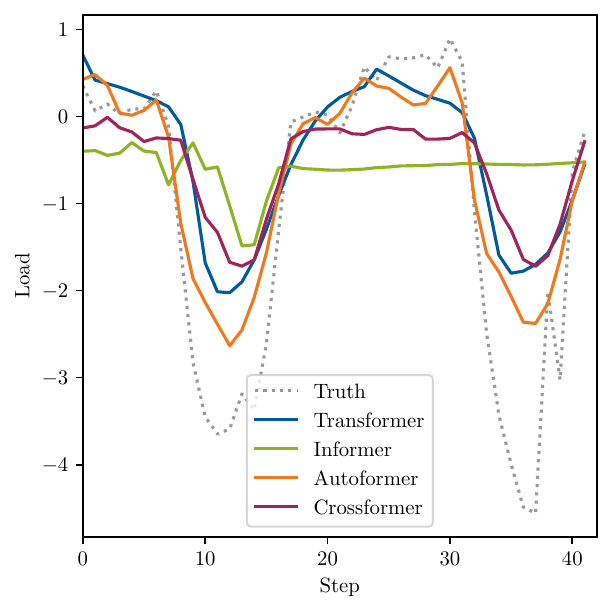}
    \caption{Loss curves for all models during training on the ENTSO-E dataset, averaged over 50 steps to reduce noise; and examples predictions on the test dataset after training.}
    \label{fig:entsoe_loss_pred}
    \Description{A figure of two images, the left one showing the averaged loss on the y-axis from 0.1 to 1.0, against the update steps during training on the x-axis going from 0 to 400. The right image is an example forecast of all models produced after training, with ground truth shown behind.}
\end{figure}

Training all models on the ENTSO-E dataset produces the averaged training loss curve shown on the left in figure \ref{fig:entsoe_loss_pred}. While not fully converged in training, the models work adequately, with a sample prediction shown on the right in figure \ref{fig:entsoe_loss_pred}.
From pruning and testing on the ENTSO-E test dataset we compile the relationship between test error and density in figure \ref{fig:entsoe}. We observe that for all models, except Crossformer, a pruning to at least 50\% without significant performance loss is possible. Around this mark, test error rises steeply with decreasing density. 

Similar to the previous experiment, we compare against smaller models. The reduced model size is defined by the largest pruning ratio without performance loss, based on results in figure \ref{fig:entsoe} we choose a reduction of 50\%. These smaller models are trained and tested on the ENTSO-E dataset, their results are shown in table \ref{tab:small_vs_big}. In all cases, the models with a higher parameter count have a lower test error, indicating that the surplus of parameters is needed to accurately learn and generalize from the training data. This higher parameter count necessarily leads to an increase in training time. Table \ref{tab:small_vs_big} demonstrates the necessity of large parameter counts during training on large datasets. The lower loss of large models persists even after pruning, as seen in figure \ref{fig:entsoe}, showing that the more compute intensive approach of training a larger model, pruning it, with possible fine-tuning can not be replaced by selecting a smaller model size to begin with.

\section{Conclusion and Future Work}
\label{sec:conclusion}

With Transformer-based models dominating the current landscape in time-series forecasting, the growing computational demand of these models urgently needs to be reined in. This is particularly relevant as demand for their deployment in real-world applications e.g., on low-resource embedded devices grows.
Pruning methods provide a viable approach to address this issue. Our study provides valuable insights into potential and pitfalls of pruning Transformer-based model for time series forecasting.
We find that while it is possible to prune models with unstructured pruning to a sparsity of at least 50\%, only models containing Fourier series decomposition retain sufficient predictive performance up to 90\% sparsity.
While structured pruning can provide actual savings in terms of operations performed, the complex, interwoven architecture of current state-of-the-art time series forecasting Transformers pose a hindering factor. Our experiments utilizing \emph{DepGraph} demonstrate this, as the algorithm fails to prune these models as desired. 

We observe that overall the performance of pruned models is strongly coupled to the dataset at hand. Although dataset size does not play a crucial role, as seen with the small Exchange dataset, the feature dimension size seems to be the relevant factor, e.g. in the Traffic and ECL dataset.
While unstructured pruning amounts to a purely academic exercise with theoretical insights due to the lack of hardware and software support, structured pruning with \emph{DepGraph} can practically reduce overall model size. This is demonstrated by our experiments, where we observe a reduction in model parameters, and consequently, in FLOPs. Unfortunately this does not amount to any notable gains in inference runtime. This result implies that for the selected Transformer-based models most of the compute time is not spent on prunable linear layer calculations, but on the time series specific parts of time series Transformers.

One major concern when it comes to Transformer-architectures for time series forecasting is that models are simply over-parameterized for most common time-series forecasting problems, resulting in overfitting on the comparably small datasets. To account for this aspect, we specifically conducted experiments with reduced model sizes compared to the originally published versions. Our results show, that indeed smaller models yield competitive prediction results on small datasets.
Likewise, we investigate model performance and pruning potential on larger datasets. Here, we observe that the larger models with more free parameters yield better performance in terms of lower MSE, even after pruning. A large parameter count is therefore necessary during training, but can be pruned afterwards.

When it comes to inference and model deployment, pruning is typically performed in conjunction with kernel compilation for corresponding hardware, to achieve actual speed-up. TensorRT~\cite{tensorrt} is an open-source library developed by NVIDIA, providing CUDA kernels to speed up deployed models. Existing models and code can be compiled to make use of TensorRT. 
To take such an inference compilation into account, we applied TensorRT to the investigated pruned models. However, the published implementations failed to compile for all models, because of specialized code that may be too complex to be retracable by the compiler. To test the speedup possibilities of TensorRT for model deployment would require completely new re-implementations of the models. The focus of our work lies on studying the effects of pruning in existing models, rather than optimized custom implementations. Hence, we refrained from this step, given that it would go way beyond the scope of this study and leave it for future work.

The insights of our work highlight the complexity and variability in the behavior of pruned models, underscoring the need for further research into time series specific models. Future work should hence explore alternative approaches, like dynamic sparse training and lottery ticket pruning, or other compression and reduction schemes, such as low-rank factorization and quantization. It is of interest how these fare in conjunction with one another. 
Understanding these dynamics will be essential as Transformers continue to grow in size and importance, necessitating efficient methods for their deployment on limited computational resources, especially in time series forecasting tasks.

\section*{Acknowledgements}

This work is supported by the German Federal Ministry of Education and Research under the 01IS22068 - EQUIPE grant. The authors gratefully acknowledge the computing time made available to them through the HAICORE@KIT partition and on the high-performance computer HoreKa at the NHR Center KIT. This center is jointly supported by the Federal Ministry of Education and Research and the state governments participating in the NHR (\url{www.nhr-verein.de/en/our-partners}).

\bibliography{literature}
\bibliographystyle{ACM-Reference-Format}
\end{document}